# CNS-Obsidian: A Neurosurgical Vision-Language Model Built From Scientific Publications

Anton Alyakin, MSE[1,2,3,†], Jaden Stryker, BS[1], Daniel Alexander Alber, BS[1,4], Jin Vivian Lee, MD[1,2,5], Karl L. Sangwon BS[1,4], Brandon Duderstadt, MSE[6], Akshay Save, MD[1], David Kurland, MD[1], Spencer Frome, BS[1,4], Shrutika Singh[1], Jeff Zhang, MSc[1,7,8], Eunice Yang, BS[1,9], Ki Yun Park, PhD[2,3], Cordelia Orillac, MD[1], Aly A. Valliani, MD[1], Sean Neifert, MD[1], Albert Liu, MD[1], Aneek Patel, MD[1], Christopher Livia, MD[1], Darryl Lau, MD[1], Ilya Laufer, MD[1], Peter A. Rozman, MD[1], Eveline Teresa Hidalgo, MD[1], Howard Riina, MD[1,10], Rui Feng, MD[11], Todd Hollon, MD[12], Yindalon Aphinyanaphongs, MD, PhD[7,8], John G. Golfinos, MD[1,13], Laura Snyder, MD[14], Eric C. Leuthardt, MD, MBA[2], Douglas Kondziolka, MD[1,15], Eric Karl Oermann, MD[1,5,10,16,17,‡]

**Affiliations:**

[1]Department of Neurological Surgery, NYU Langone Health; New York, 10016, USA.
[2]Department of Neurosurgery, Washington University in Saint Louis; Saint Louis, MO 63110, USA.
[3]Washington University School of Medicine; Saint Louis, MO 63110, USA.
[4]New York University Grossman School of Medicine; New York, NY 10016, USA.
[5]Gloabl AI Frontier Lab, New York University; New York, NY 11201, USA
[6]Nomic AI; New York; NY 10003, USA.
[7]Department of Population Health, NYU Langone Health; New York, NY 10016, USA.
[8]Division of Applied AI Technologies, NYU Langone Health; New York, NY 10016, USA.
[9]Columbia University Vagelos College of Physicians and Surgeons; New York, NY, USA.
[10]Department of Radiology, NYU Langone Health; New York, NY 10016, USA.
[11]Department of Neurosurgery, Mount Sinai Health System; New York, NY 10019, USA.
[12]Department of Neurosurgery, University of Michigan; Ann Arbor, MI 48109, USA.
[13]Department of Otolaryngology - Head and Neck Surgery, NYU Langone Health; New York, NY 10016, USA.
[14]Department of Neurosurgery, Barrow Neurological Institute; Phoenix, AZ 85013, USA.
[15]Department of Radiation Oncology, NYU Langone Health; New York, NY 10016, USA.
[16]Center for Data Science, New York University; New York, NY 10011, USA.
[17]Neuroscience Institute, NYU Langone Health; New York, NY 10016, USA.

**Author contributions:**

Conceptualization and supervision: EKO
Journal publication data collection: AA, JS, KS, DKu
Data extraction, processing, filtering, and organization: AA, JS
Data embedding, mapping, and visualizations: BD
Data conversion pipeline development: AA
Model development and training: AA
Model evaluation suite: JS
Benchmarking and ablations: AA, JS, EKO
Randomized trial user interface and web stack development: JS, EKO
Clinician onboarding to trial interface and patient data provision: AS, CO, AV, SN, AL, AP, CL, IL, DO, DK, EKO
Retrospective assessment of final diagnoses: AS, SF
Statistical analyses of trial data: AA, EKO
Manuscript figure design: AA, DAA, BD, KS, EKO
Manuscript drafting: AA, DA, JVL, EKO
Manuscript revision and final editing: All authors

[†]**Corresponding author (during review):** Department of Neurosurgery, NYU Langone Medical Center, New York University, 550 First Ave, MS 3 205, New York, NY10016, USA. Email: alyakin314@gmail.com

## Abstract

### Background and Objectives

General-purpose vision-language models (VLMs) demonstrate impressive capabilities, but their opaque training on uncurated internet data poses critical limitations for high-stakes decision-making, such as in neurosurgery. We present CNS-Obsidian, a neurosurgical VLM trained on peer-reviewed neurosurgical literature, and demonstrate its clinical utility compared with GPT-4o in a real-world setting.

### Methods

We compiled 23,984 articles from *Neurosurgery Publications* journals, yielding 78,853 figures and captions. Using GPT-4o and Claude Sonnet-3.5, we converted these image-text pairs into 263,064 training samples across three formats: instruction fine-tuning, multiple-choice questions, and differential diagnosis. We trained CNS-Obsidian, a fine-tune of the 34-billion parameter LLaVA-Next model. In a blinded, randomized deployment trial at NYU Langone Health (Aug 30–Nov 30, 2024), neurosurgeons were assigned to use either CNS-Obsidian or a HIPAA-compliant GPT-4o endpoint as a diagnostic co-pilot after patient consultations. Primary outcomes were diagnostic helpfulness and accuracy, assessed via user ratings and presence of the correct diagnosis within the VLM-provided differential, respectively.

### Results

CNS-Obsidian matched GPT-4o on synthetic questions (76.13% vs 77.54%, p=0.235), but only achieved 46.81% accuracy on human-generated questions versus GPT-4o's 65.70% ($p<10^{-15}$). In the randomized trial, 70 consultations were evaluated (32 CNS-Obsidian, 38 GPT-4o) from 959 total consults (7.3% utilization). CNS-Obsidian received positive ratings in 40.62% of cases versus 57.89% for GPT-4o (p=0.230). Both models included correct diagnosis in approximately 60% of cases (59.38% vs 65.79%, p=0.626).

### Conclusions

Domain-specific VLMs trained on curated scientific literature can approach frontier model performance in specialized medical domains despite being orders of magnitude smaller and less expensive to train. This establishes a transparent framework for scientific communities to build specialized AI models. However, low clinical utilization suggests chatbot interfaces may not align with specialist workflows, indicating need for alternative AI integration strategies.

## Introduction

Integrating generative artificial intelligence (AI) into neurosurgery may improve clinical workflows, decision-making, and patient outcomes.[1] Proprietary large language models (LLMs), such as GPT-4o and Llama-3, possess sufficient neurosurgical knowledge to pass board-like examinations[2], while ChatGPT reduces time spent writing discharge notes and operative reports.[3] While general-purpose LLMs demonstrate impressive capabilities, their opaque training on internet data may limit high-stakes neurosurgical applications.[1,4] AtlasGPT addresses some of these concerns using retrieval-augmented generation (RAG) with 250,000 pages of neurosurgical literature and outperforms generalist LLMs on board exam benchmarks.[5] However, this approach still relies on a closed-source LLM and cannot process essential visual data such as diagnostic imaging.

The emergence of vision-language models (VLMs) represents a paradigm shift in surgical AI. SurgVLM trained on approximately 1.8M surgical images across 16 surgery types, not including neurosurgery, outperformed generalist models.[6] While promising, building VLMs requires massive datasets not available in neurosurgery. We demonstrate how domain-specific models can be built using high-quality, multimodal data derived from peer-reviewed literature. Our approach addresses both the need for transparent, verifiable training data and the lack of visual reasoning capabilities in neurosurgery-specific models.

In this study, we built a vision-language training dataset from figures and text in neurosurgical articles from the *Neurosurgery Publications* journal family. We implemented a novel, three-stage curriculum learning pipeline to fine-tune a 34-billion-parameter vision-language model. Our model, CNS-Obsidian, was deployed as a diagnostic co-pilot in a web interface and compared to a state-of-the-art generalist baseline, GPT-4o; in a prospective, randomized trial, CNS-Obsidian matched GPT-4o in diagnostic performance. Our work establishes a reproducible framework for training specialty-specific VLMs using curated datasets from peer-reviewed literature. (**Fig. 1**)

## MATERIALS AND METHODS

### Study Design

This study was approved by the NYU Langone Institutional Review Board (i23-00510); the need for informed consent was waived. The protocol was reviewed and approved by the leadership of *Neurosurgery Publications* and *Wolters Kluwer*. *Self-Assessment of Neurological Surgery* (SANS) questions were used with permission from the CNS.

### Data Acquisition and Processing

We extracted data from three neurosurgical journals: *Neurosurgery* (n=21,219 articles), *Operative Neurosurgery* (n=2,618 articles), and *Neurosurgery Practice* (n=147 articles) (See **Supplemental Digital Content 2. Methods: Data Acquisition**).[7,8] We used a ResNet-based classifier (see **Supplemental Digital Content 2. Methods: Image Filtering**) to filter 78,853 figures into three categories: patient imaging, clinical visuals, and technical content.

### Knowledge Translation Framework

*Conversion Pipeline*

We used OpenAI GPT-4o[9] and Claude Sonnet-3.5[10] to convert neurosurgical figures and captions into three vision-language datasets:

1. Instruction fine-tuning (IFT): Conversational question-answer pairs. (127,076 samples, 4.2M tokens)
2. Multiple-choice (MC): Clinical vignettes with answer options (89,587 samples, 3.7M tokens)
3. Differential diagnosis (DDx): One-line case summaries with tiered diagnoses (46,401 samples, 0.4M tokens)

Models were prompted to generate task-specific user-assistant conversations based on the image, caption, and in-text mentions. Each query included four randomly sampled manual conversion examples from a set of ten, and the visual data of the target figure as a byte-string. We used structured JSON outputs and performed post-hoc manual cleaning.

## Dataset Visualization

*Data Cartography*

To visualize differences between our specialty-specific dataset and a public general medicine datasets, we embedded the text portion of the IFT dataset with a similarly generated PubMed dataset[11] into a shared two-dimensional space. We used Nomic-Embed-Text-v1[12], to convert unstructured text into numerical vectors. We applied tSNE[34] for dimensionality reduction and identified 12 largest clusters using HBDSCAN[13]. Labels were assigned by sampling texts from each cluster and querying GPT-4o to generate a unifying thematic name. Comparison of NeuroPubs with PubMed revealed gaps in broader AI datasets, such as the underrepresentation of cerebral and spinal imaging (**Fig. 2**).

## Model Architecture and Training

*Vision-Language Model Backbone*

Large Language and Visual Assistant (LLaVA)[14] is an autoregressive, multimodal VLM that extracts image features using a vision-transformer and projects them into language space, enabling image-conditioned text generation. LLaVA-Next[15] improved this design through multilayer projections and patch-based processing of high-resolution images. (**Fig. S2A**) We finetuned 34-billion parameter LLaVA-Next checkpoint on our neurosurgery-specific content.

*Three-Stage Curriculum Training*

Based on the LLaVA-Med medical curriculum training[11], we designed a three-stage fine-tuning protocol for a specialist model (**Fig. S2B**):

1) Medical alignment: we kept the language model frozen and trained only the projection layers on PubMed-based figure-caption pairs
2) Medical knowledge integration: we trained both projection and language models on general medicine PubMed-based conversations (IFT) dataset generated using GPT-4o.
3) Neurosurgical specialization: we maintained the same unfrozen components while training on our domain-specific, Neurosurgery Publications-based, and task-specific (IFT, MC, and DDx) datasets generated using GPT-4o and Claude Sonnet-3.5.

*Training Details*

The model was trained using 104 NVIDIA H100 GPUs (13 nodes, 8 GPUs each) on NYU Langone's UltraViolet high-performance computing cluster. We used PyTorch FSDP, bfloat16 precision, learning rates of 1e-3 (Stage 1) and 1e-5 (Stages 2-3), and cosine scheduling. Mini-batch size was four per GPU, with four-step gradient accumulation in Stage 1, yielding effective batch sizes of 1664 for Stage 1 and 416 for Stages 2 and 3. Similar to LLaVA-Med[11], we froze the vision encoder. We maintained paper-level data splits, allocating 95% of articles for training, and 2.5% each for validation and test sets. Validation set was used to optimize training and guide model selection, while the held-out MCQ test set was only used to evaluate final models.

*Training Length*

We refer to checkpoints of CNS-Obsidian as [<#_of_Stage_1_epochs>, <#_of_Stage_2_epochs>, <#_of_Stage_3_epochs>] for brevity and convenience. We recreated LLaVA-Med's framework with LLaVA-Next architecture, with one epoch of Stage 1 training and three epochs of Stage 2 training, yielding our LLaVA-Next-Med-OLAB [1, 3, 0]. We initially trained this model on three epochs of GPT-Only Stage 3 datasets, yielding [1, 3, 3]. Through extensive experiments and ablation studies (see **Fig. S3-S5** for results; also see **Supplemental Digital Content 3. Results: Ablations.**), we found that the standard training duration of one epoch for alignment and three epochs for medical and neurosurgical fine-tuning was insufficient for a model of this scale (**Fig. S3-S4**). Our best performing model ended up being extensively trained [5, 10, 10]. Throughout our training length experiments we only used GPT-sourced data for training. We also experimented with training this model on Claude and GPT-based data together, trying both the constraint for epoch number and compute amount **(Fig. S5**). Wall-clock training time per epoch was 3.5 hours for Stage 1, 30 minutes for Stage 2, and 1 hour for Stage 3 using only one GPT-4o data and 2 hours using both sources.

## Generative AI Evaluation

*Benchmarking*

We compared LLaVA-Next-Med-OLAB and CNS-Obsidian to multiple baselines (LLaVA-Med, LLaVA-Next, OpenAI GPT-4o, and Anthropic Claude Sonnet-3.5). For LLaVA-Med, LLaVA-Next, as well as our models we used vLLM for inference. For GPT-4o and Sonnet-3.5 evaluations, we made API calls to the publicly available checkpoints. We used LLaMA-3-70B to parse responses and extract single-letter answers.

*Synthetic Benchmarks*

Synthetic benchmarks consisted of 1,282 GPT-generated and 1,239 Claude-generated MCQs from the held-out test data. Paper-level splitting ensured that these test questions did not overlap with questions or figures seen during training.

*Human-Made Benchmarks*

Human-made benchmarks consisted of 950 image-associated questions from the SANS question bank, developed by the CNS for neurosurgery residents. These served as our primary benchmark.

## Randomized Trial

In a controlled comparison of diagnostic accuracy and utility, we randomized neurosurgery consults to either CNS-Obsidian or a PHI-safe version of GPT-4o. To facilitate this, we developed a full-stack application for automated blinding and randomization. (See **Fig. S6**, **Supplemental Digital Content 2. Methods: Full-Stack Web Application**). Data was collected for three months, from August 30th, 2024 to November 30th, 2024, at NYU Langone Health, Tisch Hospital for all neurosurgery consultations. All trainees and faculty in the department were invited to participate via email. Participation was voluntary and no compensation was provided. Participants were instructed to interact with the software after finishing an encounter with a patient.

## Outcomes

*Diagnostic Helpfulness*

Clinicians submitted images with clinical one-liners. Models were prompted to generate a differential diagnosis in their first response, which clinicians rated as clinically helpful ("thumbs up") or not ("thumbs down"). Clinicians could optionally continue the conversation, and follow-up chats involved next-step clinical decision-making queries such as diagnostic workup and best management. The primary outcome of the study was the diagnostic helpfulness of the first response, defined as the frequency of “thumbs up”.

*Diagnostic Accuracy*

Patient identifiers were recorded but omitted from model inputs. Ground truth diagnoses were retrospectively retrieved and compared against model-generated differential diagnoses. Accuracy of the differential diagnosis, measured as the proportion of cases where the true diagnosis appeared in the list of differentials served as an additional primary outcome. Evaluation involved a two-step automated process:

- GPT-4o extracted individual diagnoses from the unstructured model outputs into a structured list.
- GPT-4o determined whether the true diagnosis was included in this list.

During analysis, GPT-4o's differentials were observed to be longer and less specific compared to CNS-Obsidian. To address this, we introduced an ad-hoc length-adjusted accuracy metric: the correct diagnosis rate per differential, measured as the proportion of total items in the list with a correct diagnosis.

*User Engagement*

Secondary outcomes included the conversation length and the clinical helpfulness of the follow-up chats. Conversation length was measured as the total number of user messages in the chat per conversation (baseline of 1). Clinical helpfulness of follow-up chats was measured as a binary outcome with upvote or downvote by the user. We also recorded the total number of clinician interactions and total number of neurosurgery consults during the study period.

**Statistical Analysis**

For the MCQ performance evaluations, two-sided Fisher's exact tests were performed to assess differences in proportions (accuracy) between models across benchmarking comparisons. The randomized trial was designed to compare differences in subjective (diagnostic helpfulness, clinical helpfulness) and objective (diagnostic accuracy, length-adjusted diagnostic accuracy, conversation length) outcomes between CNS-Obsidian and GPT-4o. Two-sided Fisher's exact tests were performed to assess differences in proportions (all outcomes except conversation length) between models. Two-sided Mann-Whitney *U* test was performed to assess differences in distributions (conversation length) between models.

## RESULTS

### Benchmark performance

CNS-Obsidian matched GPT-4o's MCQ accuracy on the held-out GPT-generated MCQs (p=0.2347) (**Fig. 3C**) and also outperformed both GPT-4o and Claude Sonnet-3.5 on Claude-generated MCQs (p=0.0011 and 0.0004, respectively), despite being exclusively trained on the GPT-generated data (**Fig. 3D**). Zero-shot accuracy on the human-generated CNS-SANS questions improved from 39.81% to 46.81%, but still lagged behind GPT-4o (65.70%, $p<10^{-15}$) (**Fig. 3E**). Adding Claude-generated data to the training set improved accuracy on Claude-generated MCQs (p=0.0427) but not performance on GPT-generated MCQs (p=1.000) or SANS (p=0.5193) (**Fig. 3C-E**).

### CNS-Obsidian as a clinical diagnostic co-pilot

During the 92-day trial, a total of 70 chats were generated (0.75/day; 38 randomized to GPT-4o, 32 randomized to CNS-Obsidian) (**Fig. 4, Fig. 5A, Fig. S7, Videos 1 and 2**) and 959 total consults seen by neurosurgery (10.42/day), suggesting meaningful but modest AI engagement (7.3% utilization rate). For our primary endpoint—diagnostic helpfulness—CNS-Obsidian achieved a 40.62% upvote rate versus 57.89% for GPT-4o (p=0.2301) (**Fig. 5B**). For the secondary endpoint—diagnostic accuracy—CNS-Obsidian included the correct diagnosis at the time of consultation in 59.38% of cases versus GPT-4o did in 65.79% (p=0.6259) (**Fig. 5C**). GPT-4o tended toward broader, lengthier differentials (**Fig. 5A**). Adjusting for response length, CNS-Obsidian trended toward higher accuracy per differential (16.88% vs. 10.69%, p=0.0809) (**Fig. 5D**). Users were encouraged to ask follow-up questions such as further diagnostic work-up or suggested management. For our assessment of user engagement, average conversation length was 2.50 (CNS-Obsidian) versus 1.79 messages (GPT-4o, p=0.6655) (**Fig 5E**), with users rating the helpfulness of follow-up conversation lower for CNS-Obsidian than for GPT-4o (upvote rate 25.00% vs. 40.00%, p=0.2094) (**Fig. 5F**).

## DISCUSSION

Previous attempts at using scientific publications to train vision-language models have mostly produced non-generative contrastive learning models.[16–19] Some generative works have explored converting figure captions to simple visual question-answering datasets using GPT-4o.[11,20] Independent parallel efforts[21] trained LLaVA-Next on a fusion of PubMed[11] and general internet data[15], yielding a different version of LLaVA-Next-Med. Importantly, all of these works use the publicly available PubMedCentral. Separately, VLMs have been fine-tuned on scraped medical textbooks,[22] but without the IFT they are sometimes outperformed even by baseline models. CNS-Obsidian is the first VLM trained on specialty-specific peer-reviewed literature. Securing explicit permissions from the publishers demonstrates a potential future avenue of ethical and productive symbiotic relationship between AI and scientific publications. Our work is unique in its data source, experimentation extensiveness, and the goal of serving as a clinical co-pilot.

CNS-Obsidian performed comparably to GPT-4o and Claude Sonnet-3.5 on neurosurgical MCQs, reinforcing the importance of the in-domain data. Unlike prior approaches which relied on prompt engineering[23] or RAG[5] we directly fine-tuned our model on visual differential diagnoses cases. Since high-quality medical information is scarce in public datasets[11] and uncurated internet sources such as Twitter[24], scientific publications offer an immediate solution, particularly for multimodal paired image-text context, which is rare outside of curated sources. To ensure rigorous evaluation, we performed a paper-level data split, holding out entire articles from training to minimize potential data contamination. In contrast, GPT-4o's and Claude Sonnet-3.5's training datasets remain unknown, with published instances of data leakage falsely inflating benchmark performance[25,26]. Given the widespread prior use of SANS in LLM benchmarking[2,27–29], partial prior exposure in frontier models remains a plausible contributing factor.

We conducted the first blinded randomized trial of VLM chatbots in a clinical setting. Unlike prior studies using simulated scenarios[23,30], our trial captured the complexities of actual patient care. Neurosurgical residents and attendings at a major academic center used a web-based chat interface to interact with a VLM, providing post-encounter, but real-time differential diagnoses. CNS-Obsidian and GPT-4o were perceived as helpful at the point of care nearly half the time, with similar engagement levels. Our clinical deployment was conducted using a checkpoint obtained on August 30th, 2024, while subsequent training improvements were realized after the RCT was underway. The full point-of-care potential of our model may be underestimated.

A key finding was the low utilization rate of the chatbot by neurosurgical participants. Specialists likely prefer workload automation over decision support. The chatbot interface itself may be a barrier: taking photos and typing responses may be an unnecessary burden for specialists but could be more useful for generalists practitioners, as the broader scope of their practice makes AI-driven insights more valuable. Alternative interfaces beyond chatbots, such as passive AI-assisted workflows (e.g. ambient dictation), may be more aligned with specialist needs.

Interestingly, the delta between CNS-Obsidian and GPT-4o in subjective diagnostic helpfulness (-17.27%) was notably greater than that in objective diagnostic accuracy (-6.41%). Our training dataset consisted of structured differential diagnoses without explanations or conversational framing, reflecting our hypothesis that specialists prefer direct, focused answers. In practice, GPT-4o's more expansive, explanatory output style was perceived as more helpful. It appears that when specialists do engage with chatbots, they tend to value interpretability (reasoning transparency) and broader contextual thinking, rather than diagnostic focus.

**Limitations**

Our study focused on fully autoregressive models, excluding cross-attention-based architectures[31,32]. Future studies could explore prompt-based techniques like RAG or chain-of-thought prompting to improve performance. We also did not experiment with system prompting of the models. Second, to ensure compliance with copyright laws and AI-training ethical standards, our dataset was limited to a single journal family, *Neurosurgery Publications*, rather than the full scope of neurosurgical literature. Third, though it remains an open question whether models trained on copyrighted data transformed by other LLMs constitute the same copyright, we erred on the side of caution and respect to the publisher, keeping CNS-Obsidian's weights private. Lastly, our blinded randomized trial had a low participation rate and was conducted at a single institution, limiting generalizability and emphasizing the need for multicenter studies. The low participation rate raises questions about improving human-AI interactions in medicine, echoing findings from a recent study where surgeons in Greece had more negative perceptions of a GPT-4-based chatbot than other physicians.[33]

## CONCLUSION

Scientific literature can be used to build powerful, domain-specific AI models. Despite being orders of magnitude smaller than GPT-4o, CNS-Obsidian achieved comparable real-world diagnostic accuracy to GPT-4o (59.38% vs 65.79%, p=0.626) while providing full data auditability, computational efficiency for local deployment, and institutional customizability—advantages that address critical concerns around transparency and data sovereignty in clinical AI.

**Acknowledgements:** We appreciate the informal input from mentors, colleagues, and lab members of OLAB and Leuthardt Lab not individually acknowledged. We thank Michael Constantino, Ali Siavosh-Haghighi, Kevin Yie, and the rest of the NYU Langone High-Performance Computing (HPC) Team, who supported the computing resources fundamental to our work. Lastly, we thank the NYULH Predictive Analytics Unit for their teamwork and collaboration in making AI technologies a reality at NYULH. We also want to thank the *Congress of Neurological Surgeons* and Wolters Kluwer who run and publish *Neurosurgery Publications*, respectively. We would also like to thank the Managing Editor of *Neurosurgery Publications*, Brandon Fiedor, for his tireless work in promoting medical and scientific scholarship.

**Disclosures:** E.K.O. reports consulting with Sofinnova Partners and Google, income from Merck & Co. and Mirati Therapeutics, employment in Eikon Therapeutics, and equity in Delvi, MarchAI, Artisight. R.F., T.C.H., and E.K.O. are editorial board members of Neurosurgery Publications. D.K. is the editor in chief of Neurosurgery Publications. B.D. is a co-founder and CEO of Nomic AI.

**Funding:**
National Cancer Institute's Early Surgeon Scientist Program grant 3P30CA016087-41S1 (EKO)
W.M. Keck Foundation (EKO)
Institute for Information & communications Technology Promotion(IITP) grant funded by the Korea government(MSIT) (No. RS-2019-II190075 Artificial Intelligence Graduate School Program (KAIST); No. RS-2024-00509279, Global AI Frontier Lab). (JVL, EKO)

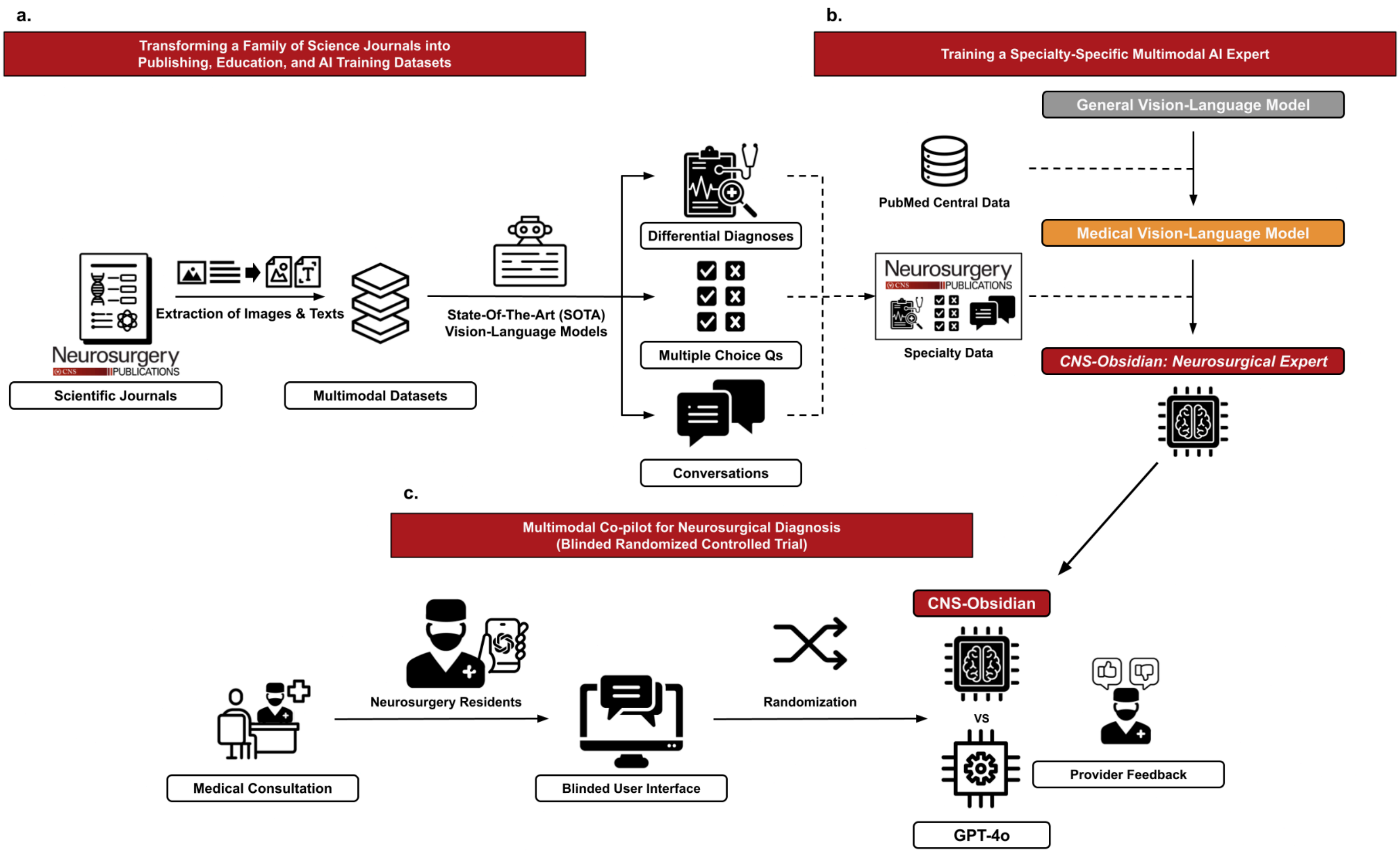

**Fig. 1. Overview of key contributions.** (**A**) We developed a pipeline for the acquisition, extraction, and filtering of figures, captions, and in-text mentions from a family of biomedical journals. We converted this data from unstructured biomedical texts and images into publication-, education-, and task-specific AI training datasets. (**B**) We trained CNS-Obsidian, a 34B parameter autoregressive vision-language model, to be tailored to domain-specific needs for neurosurgery by implementing a novel training step designed to specifically entrain capabilities in differential diagnosis while maintaining the ability to converse and answer questions. (**C**) We conducted a blinded, randomized, controlled trial comparing CNS Obsidian to GPT-4o as diagnostic copilots on a busy inpatient surgical service.

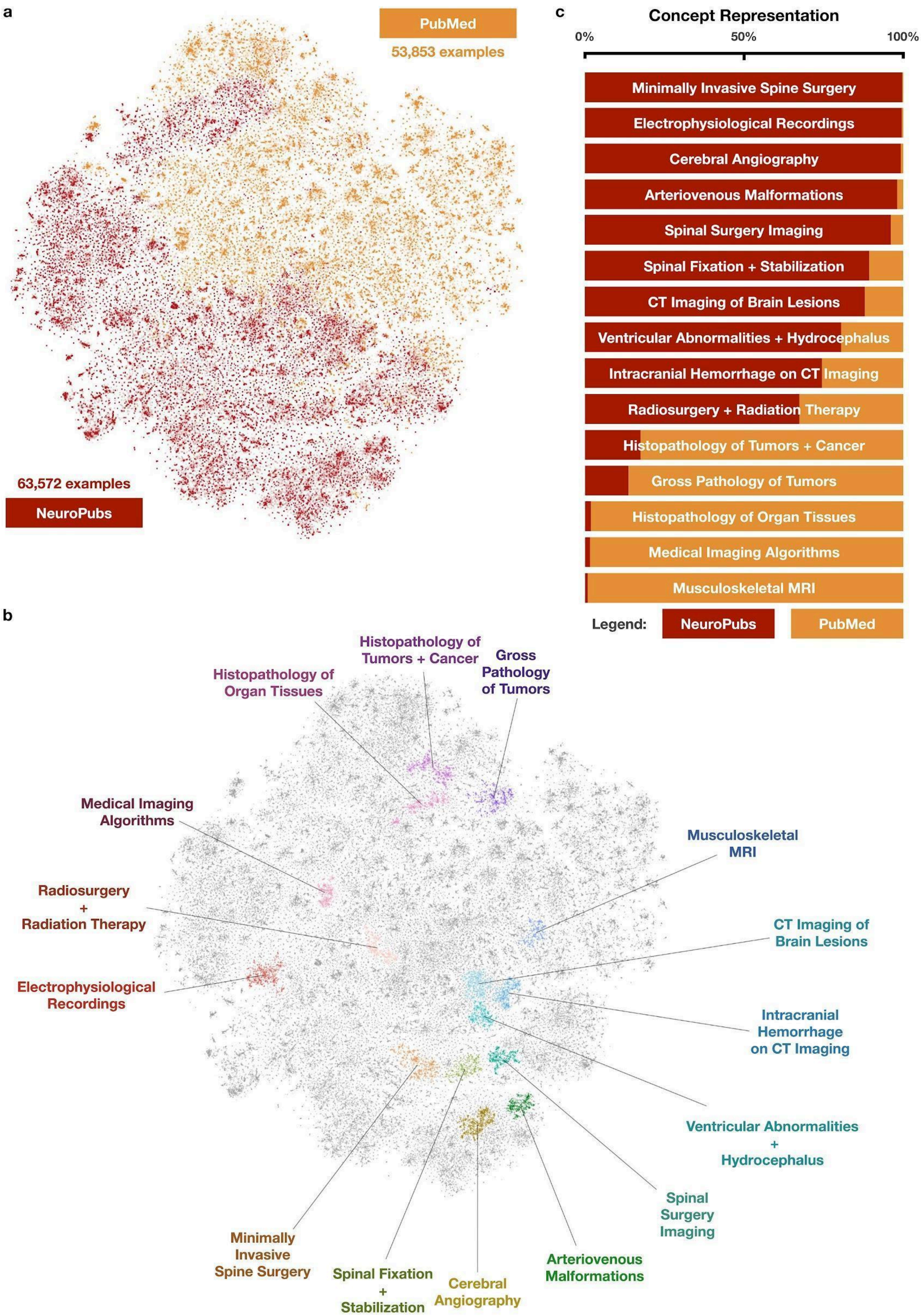

**Fig. 2. Visualization of concepts represented in scientific datasets.** (**A**) A joint embedding of the PubMed-based and Neurosurgery-based instruction fine-tuning datasets. (**B**) The 15 largest hierarchical clusters of the joint embedding. **(C**) Comparison of the percent composition of the 15 largest clusters in the datasets. NeuroPubs is particularly dense in the highly-specialized neurosurgical topics.

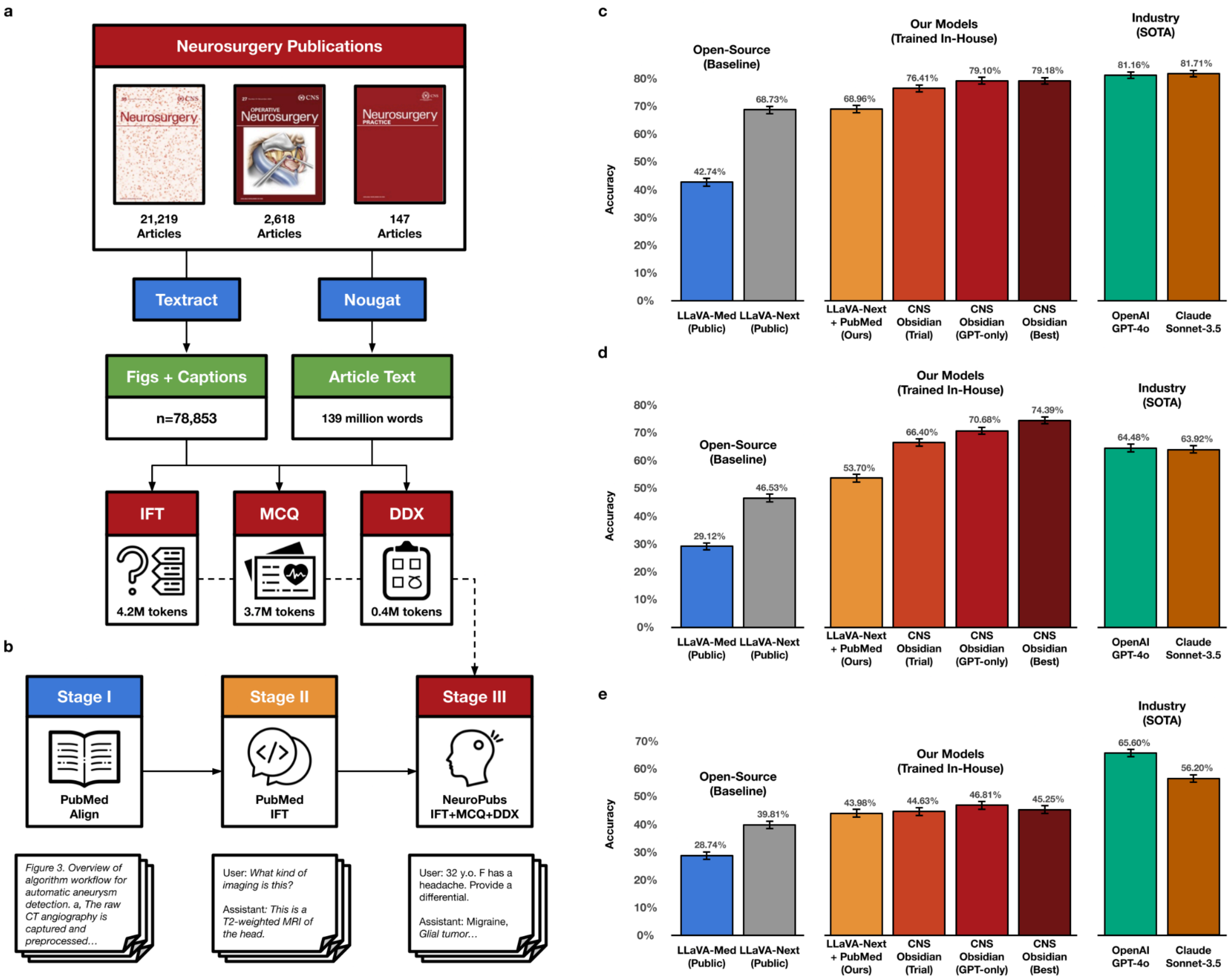

**Fig. 3. Dataset, curriculum training, and benchmark performance.** (**A**) Acquisition, processing, and knowledge translation pipeline for converting our specialty journal into an AI knowledge database (139 million words, 78,853 captioned figures) and subsequently into task-specific vision-language datasets. (**B**) Our approach includes a novel stage in training a medical specialty-specific (neurosurgical) VLM and to a downstream task (diagnosis) while preserving the conversation and question-answering capabilities of general models like ChatGPT. (**C**) MCQ performance on GPT-generated questions (n=1,282 questions) (**D**) MCQ performance on Claude-generated questions (n=1,239 questions) (**E**) MCQ performance on one-shot board-style questions (n=950 questions).
Note: Two-sided Fisher's exact tests were performed to assess differences in proportions (% accuracy) between models across benchmarking comparisons. Error bars represent standard error in all graphs.

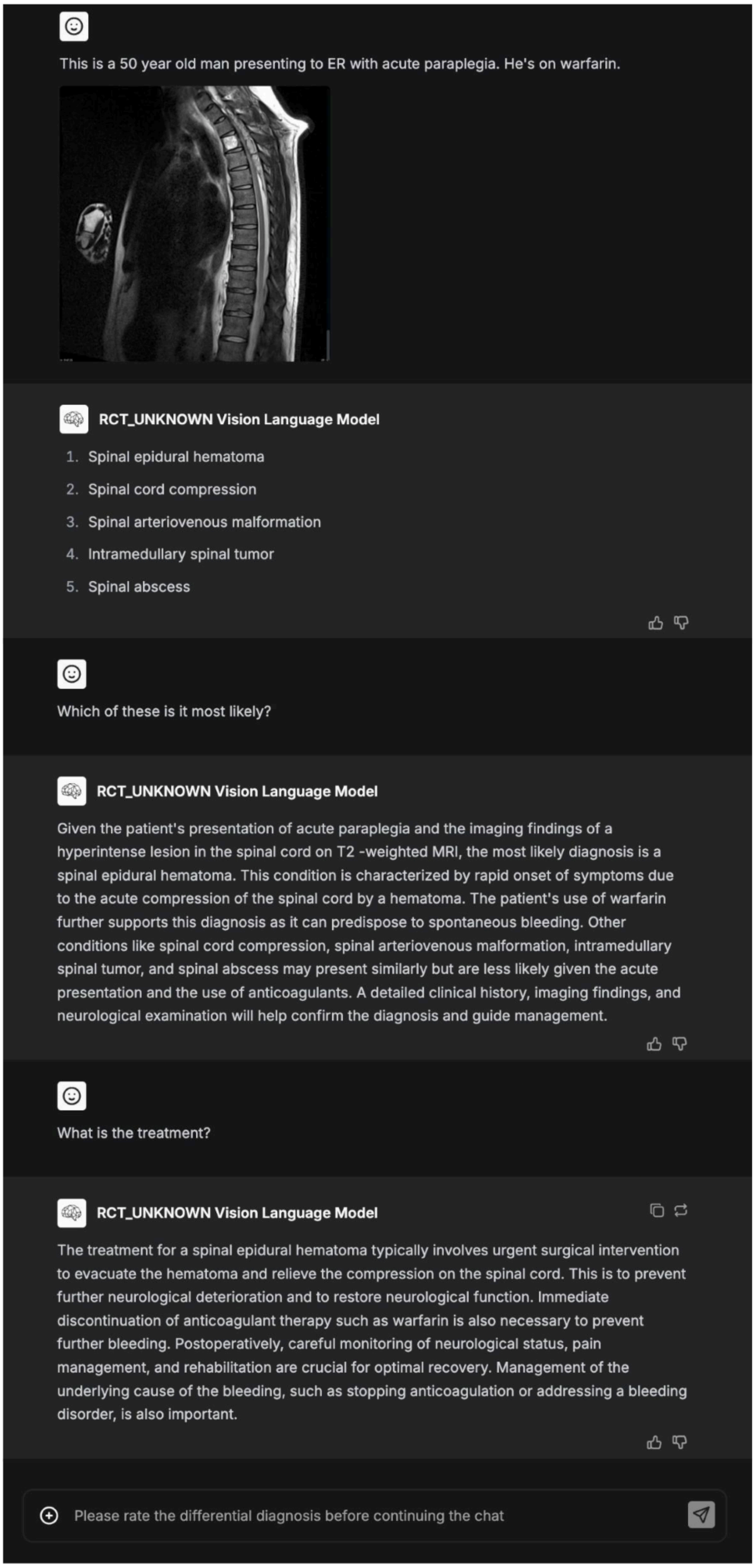

**Fig. 4. An example patient submitted to the study interface.** Identifying details anonymized. The model interacted with is CNS-Obsidian-base (version actually used in the trial). For more examples of a base and final model see **Video 1** and **Video 2**, respectively.

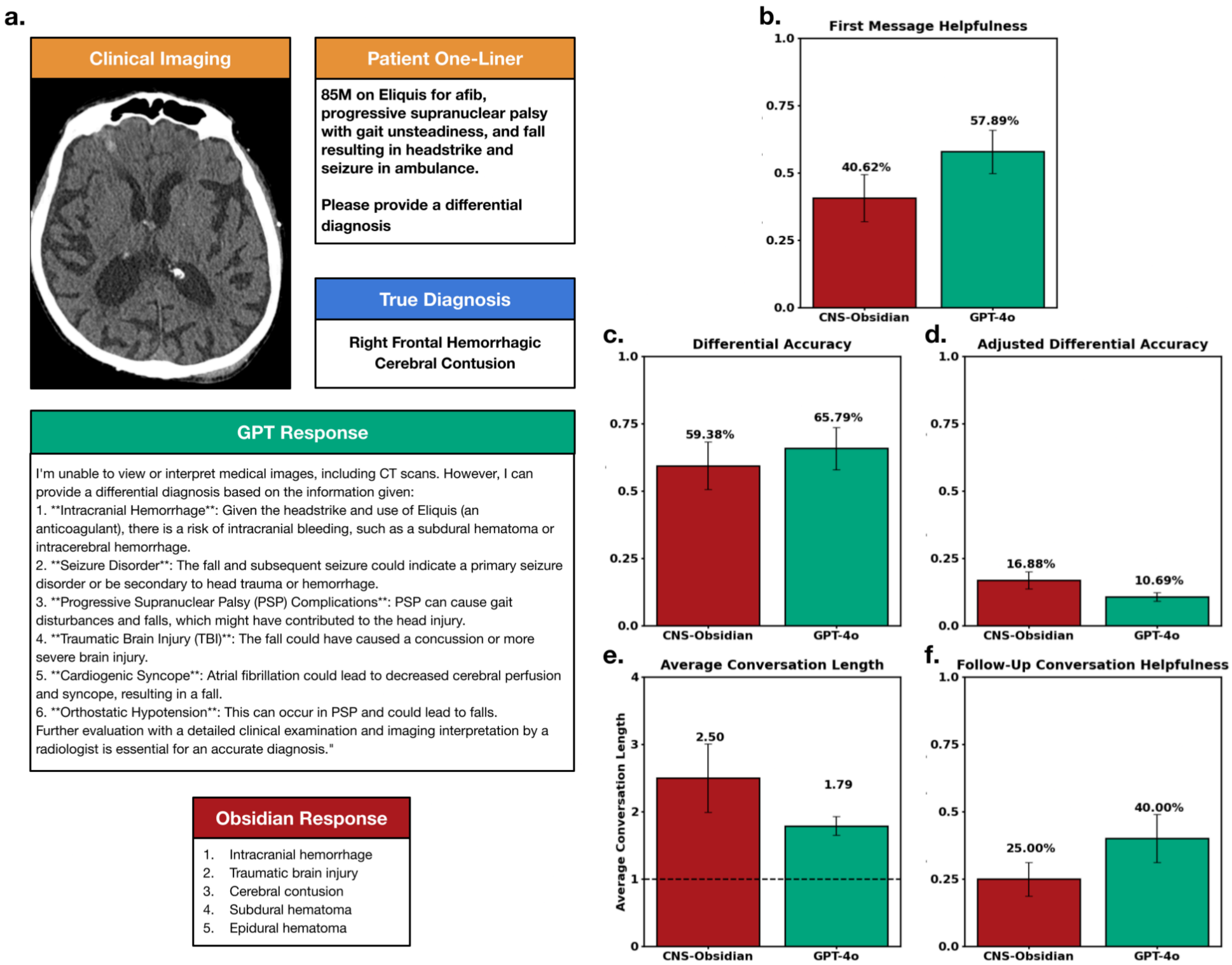

**Fig. 5. Randomized trial results.** (**A**) A visual depiction of the workflow with a representative patient encounter. During the three-month trial, users interact with a web-based chat interface to interact with the VLM for differential diagnosis assistance (see **Fig. S9** for interface details). The conversation is initialized by a neurosurgical resident who provides an image and a text input for evaluation (see **Fig. S10** for interaction details). A single representative planar image (e.g., from X-ray, CT, MRI, or digitally captured photos of the pathology) is selected at the discretion of the neurosurgery trainee, as well as a brief one-liner statement summarizing pertinent elements of the patient history (e.g., age, sex, past medical history, medications, presenting symptoms). Upon prompting, GPT or CNS-Obsidian provides a text response with a tiered differential diagnosis. Outputs from GPT, counterfactual CNS-Obsidian output, and the actual diagnosis are shown. Compared to our CNS-Obsidian, GPT generated a broader but less specific set of diagnoses in a more conversational tone. (**B**) Diagnostic helpfulness of differential (user-rated) (**C**) Diagnostic accuracy of differential (AI-rated) (**D**) Diagnostic accuracy, adjusted for response length (**E**) Conversation length (**F**) Clinical helpfulness of follow-up chats (user-rated).
Note: For panels **B-E**, data represents n=32 for CNS-Obsidian and n=38 for GPT-4o. For panel **F**, data represents n=48 for CNS-Obsidian and n=30 for GPT-4o. Two-sided Fisher's exact tests were performed to assess differences in proportions (% upvote rate) between models in panels B, C, and F. Two-sided Mann-Whitney U tests were performed to assess differences in distributions (adjusted accuracy and conversation length) between models in panels D and E. Error bars represent standard error in all graphs.

**Video 1. Interaction with CNS-Obsidian-Base-Trial: Stroke case.**
https://alyakin314.github.io/assets/converted-obsidian-base-stroke.mp4

**Video 2. Interaction with CNS-Obsidian-Final-Both: Subarachnoid hemorrhage case.**
https://alyakin314.github.io/assets/converted-obsidian-final-both-sah.mp4

**Supplemental Digital Content 1. 6 Figures.**
- Fig. S1. Dataset filtering and knowledge translation pipeline.
- Fig. S2. Training a specialty vision-language model.
- Fig. S3. Ablation studies of three-stage training using GPT-generated evaluations.
- Fig. S4. Ablation studies of three-stage training using Claude-generated evaluations.
- Fig. S5. Ablation experiments for data mixtures.
- Fig. S6. Randomized controlled trial user interface.

**Supplemental Digital Content 2. Methods**
- Data Availability
- Code Availability
- Data Acquisition
- Image Filtering
- Full Stack Web Application

**Supplemental Digital Content 3. Results**
- Ablations

# Supplementary Figures

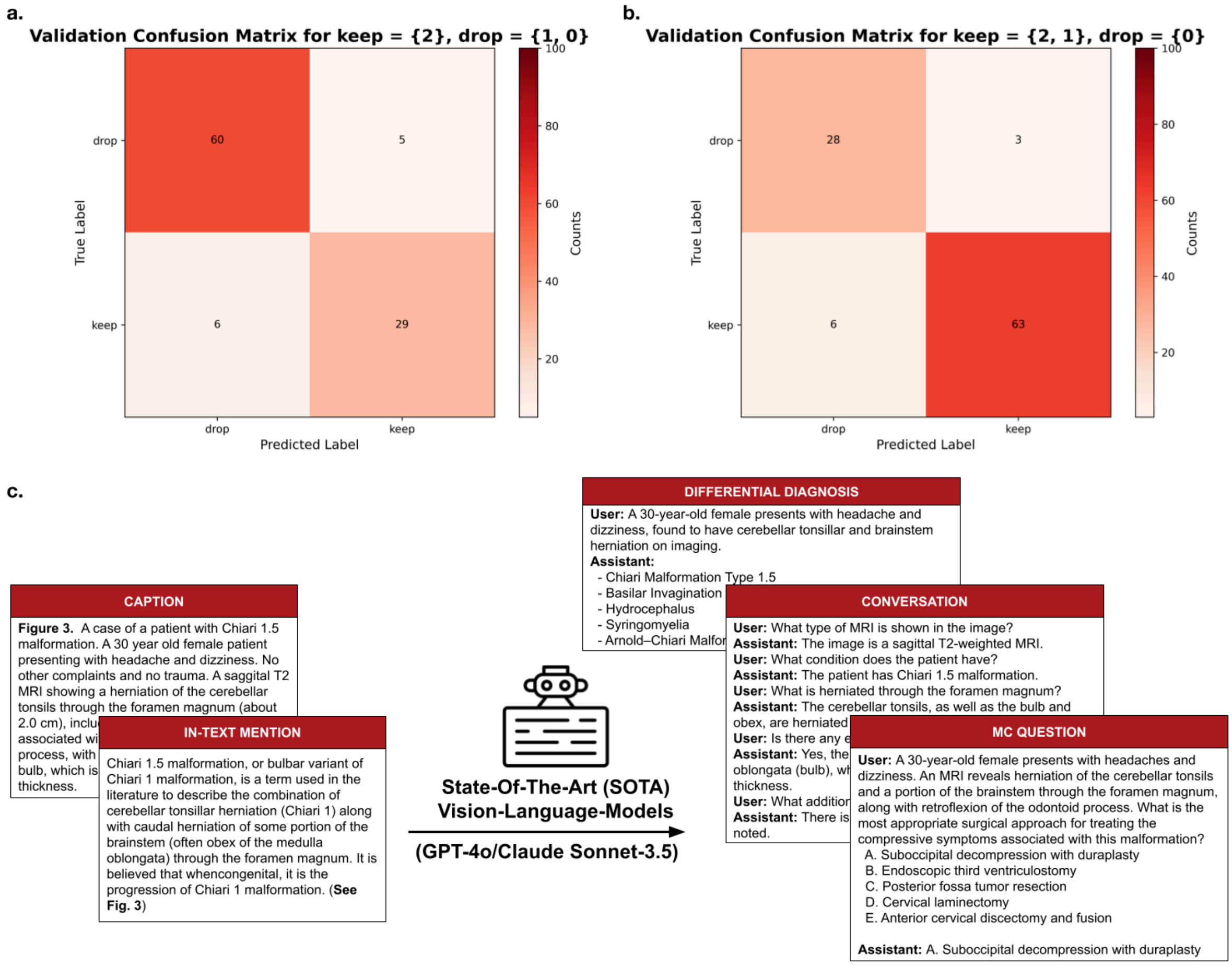

**Fig. S1. Dataset filtering and knowledge translation pipeline. (A)** Validation confusion matrix for the image content classification system when retaining Class 2 images (medical imaging: CT, MRI, X-ray, angiography) and filtering out Classes 1 (clinical visuals: surgical fields, microscopy, anatomical drawings) and 0 (technical content: flowcharts, survival curves, tables). Used to create the differential diagnosis dataset. **(B)** Validation confusion matrix for the classifier when retaining Classes 1 and 2 combined while filtering out Class 0. Used to create the multiple choice dataset. For both (A) and (B), the classifiers were based on ResNet-50 feature extraction and a linear classifier trained on 400 manually annotated images. Confusion matrices demonstrate validation performance on the held out 100 manually labeled images. **(C)** Knowledge translation pipeline demonstrated through conversion of neurosurgical data into training datasets. The example illustrates the automated conversion of specialized knowledge (figure, caption, in-text mention) into three distinct data formats: 1) natural instructional dialogue between user and assistant, 2) clinical vignette with multiple-choice options and detailed explanation, and 3) concise one-liner with prioritized differential diagnoses. The pipeline employs large language models (GPT-4o here) with few-shot in-context learning (four examples for each task randomly selected from pools of 10) to generate consistently formatted outputs while preserving diagnostic accuracy and educational value. Image: case courtesy of Rodrigo Dias Duarte, Radiopaedia.org, rID: 50409. Caption and in-text mention written based on the case information provided on Radiopaedia.org.

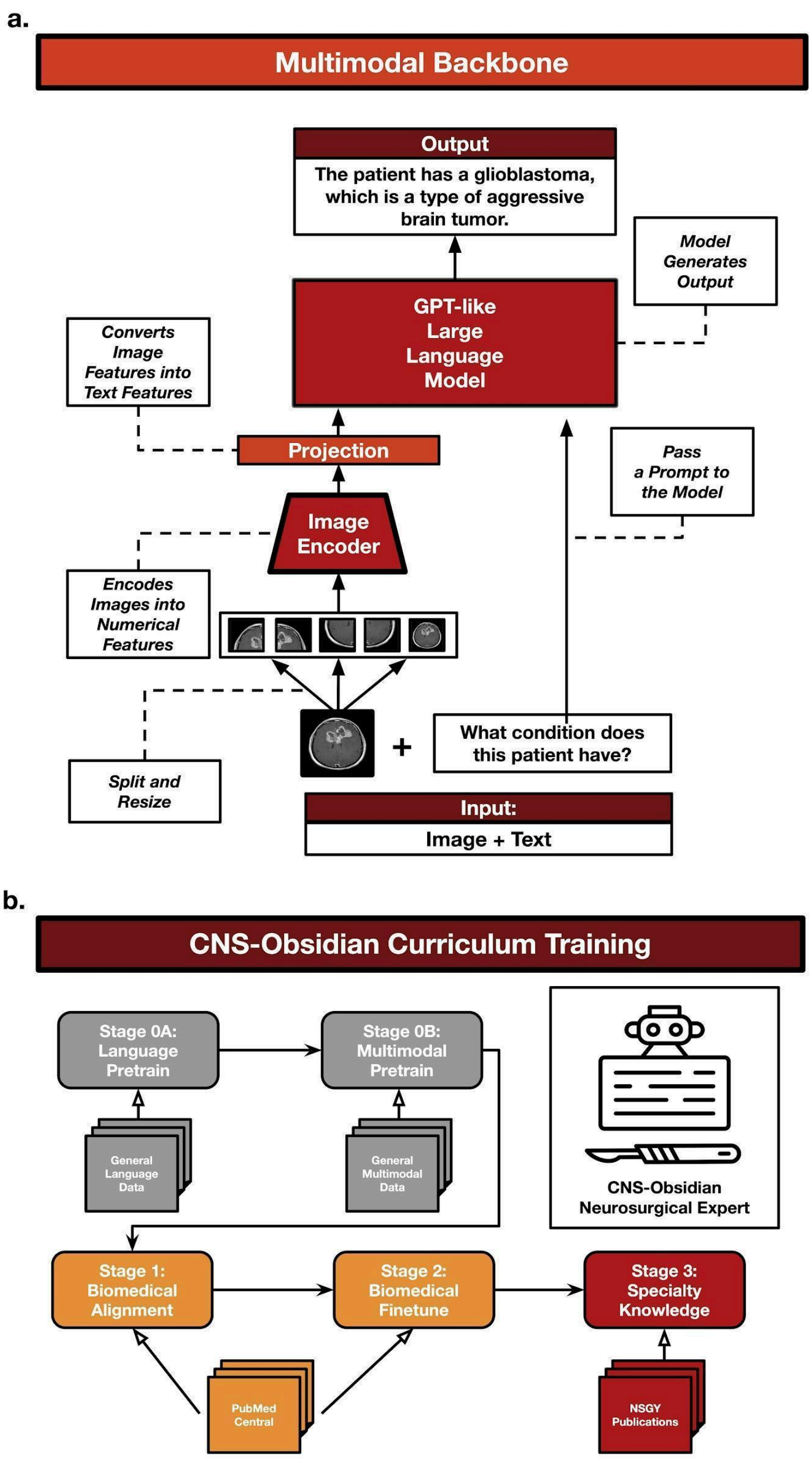

**Fig. S2. Training a specialty vision-language model.** (**A**) LLaVA-Next is a vision-language model that combines two modalities by slicing the high-resolution image into small patches and embedding them together with a resized full image. It then projects the visual features into the text space, and uses a pre-trained autoregressive model to generate output conditioned on both the image and the prompt. (**B**) A specialist model is trained in a stages, beginning with general language and multimodal pretraining (Stages 0A and 0B), followed by general medical alignment and finetuning (Stages 1 and 2), and culminating in specialty knowledge integration (Stage 3) using NeuroPubs. This curriculum-based approach was used to create CNS-Obsidian, a neurosurgical expert system capable of interpreting medical imaging and providing specialized diagnostic assessments.

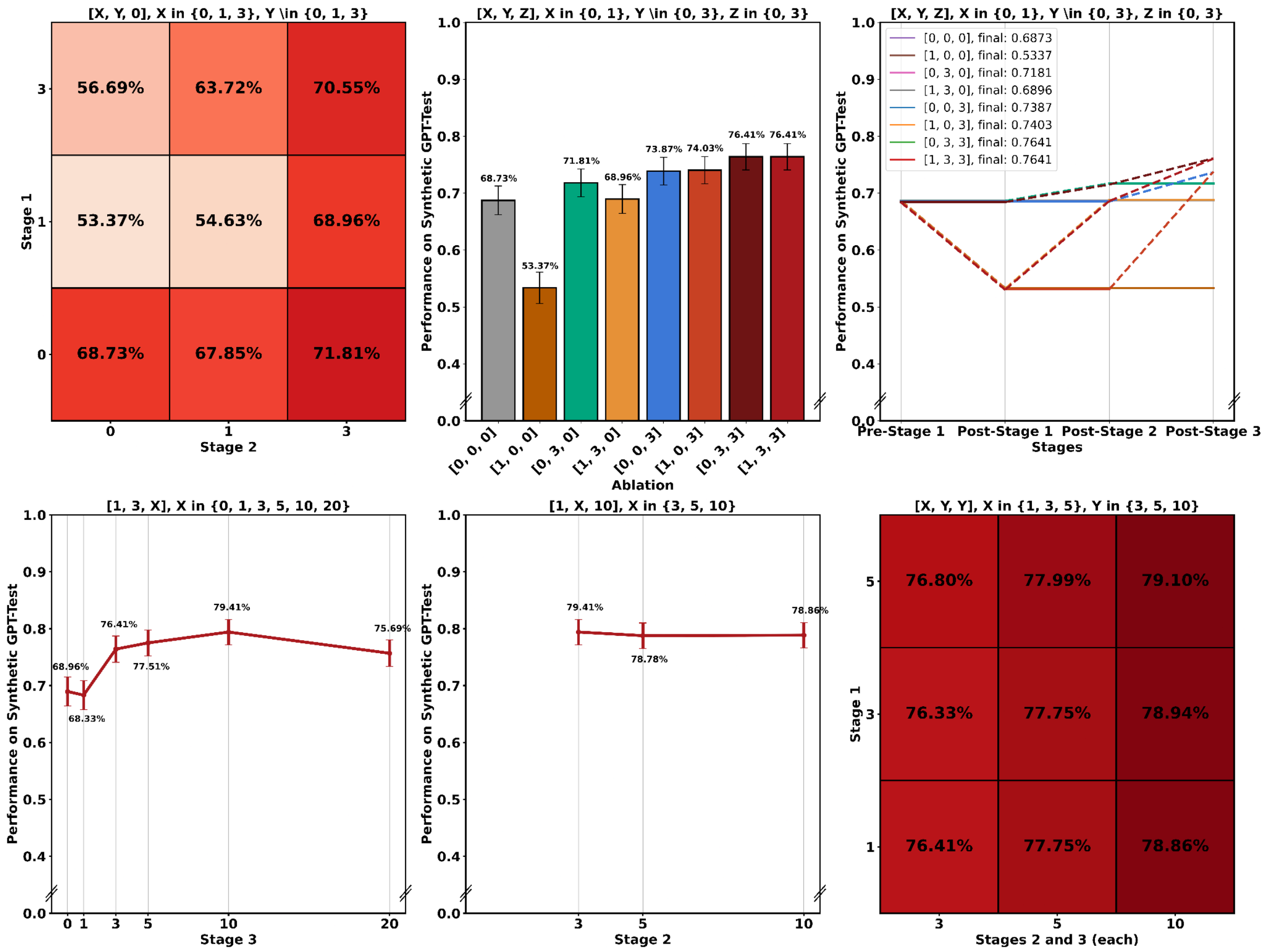

**Fig. S3. Ablation studies of three-stage training using GPT-generated evaluations.** Configuration *[X, Y, Z]* denotes number of epochs in Stages 1, 2, and 3. (**A**) Impact of Stage 1 (alignment) and Stage 2 (general fine-tuning) epochs on model accuracy using GPT tgenerated MCQs dataset (n=1,282), with Stage 3 fixed at 0. Darker red indicates higher accuracy. Baseline *[0, 0, 0]* achieves 68.73%. Alignment-only training shows performance degradation. (**B**) Performance comparison across configurations demonstrating each training stage's contribution. Error bars represent standard error. The full three-stage model *[1, 3, 3]* achieves 76.41% accuracy, with Stage 3 contributing the largest improvement (+7.45%, $p<0.0001$). (**C**) Temporal evolution of model performance across training stages for different configurations Solid lines represent measurements, dashed lines show interpolated trajectories. (**D**) Optimization of Stage 3 (task-specific fine-tuning) duration using configuration *[1, 3, X]*, where X varies from 0 to 20 epochs. Performance exhibits monotonic improvement until peaking at X = 10 epochs (79.48%). (**E**) Impact of Stage 2 duration on model performance using configuration *[1, X, 10]*, where X varies from 3 to 10 epochs. Performance remains stable across durations, *[1, 3, 10]* achieving optimal performance (79.41%) on GPT generated MCQs and selected as one of final configuration candidates. (**F**) Joint optimization of Stage 1 and Stage 2/3 durations (configurations *[X, Y, Y]*). Longer fine-tuning stages show performance improvements. *[5, 10, 10]* selected as one of the final configurations.

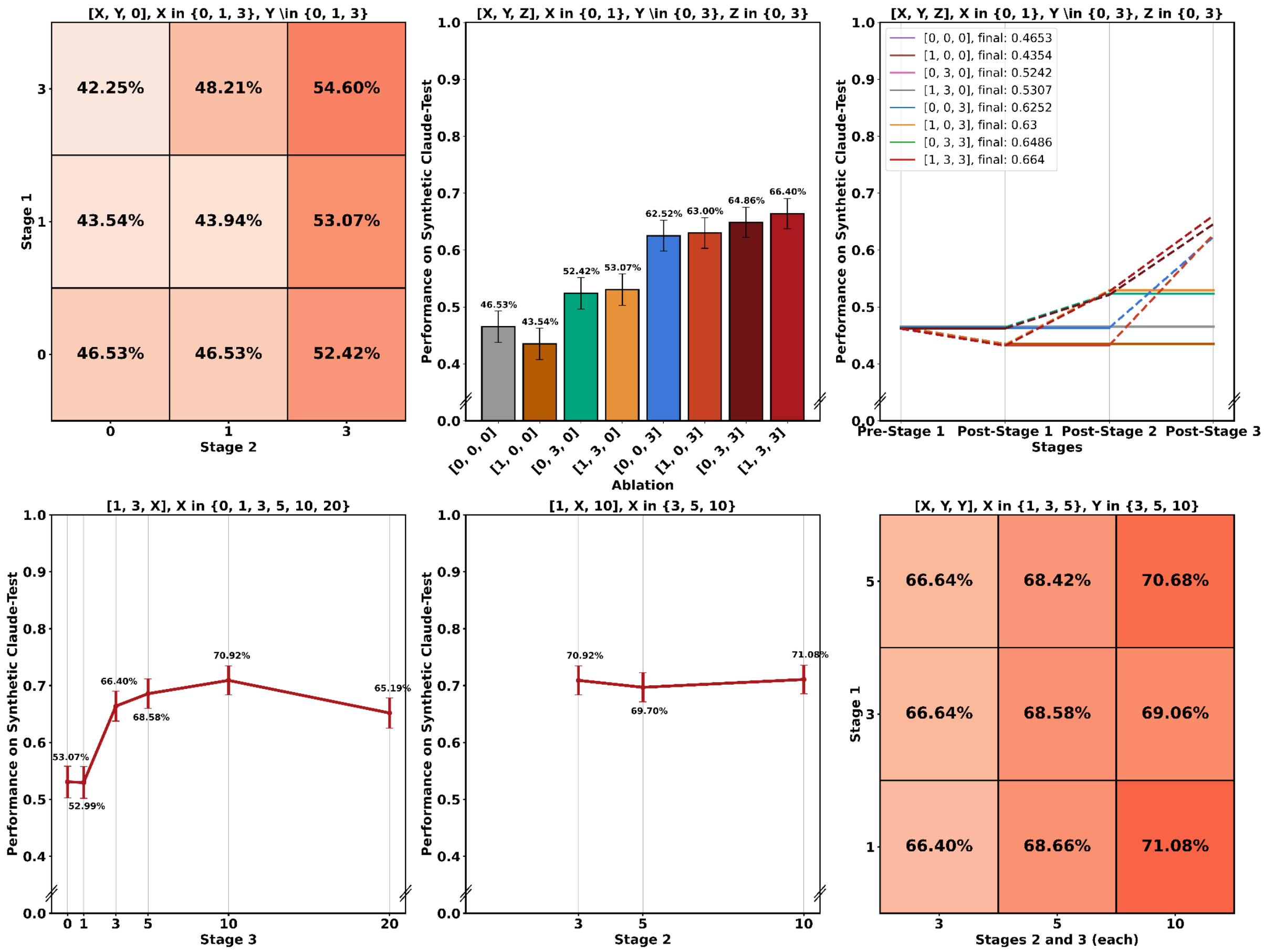

**Fig. S4. Ablation studies of three-stage training using Claude-generated evaluations.** Configuration *[X, Y, Z]* denotes number of epochs in Stages 1, 2, and 3. (**A**) Impact of Stage 1 (alignment) and Stage 2 (general fine-tuning) epochs on model accuracy using Claude generated MCQs dataset (n=1,239), with Stage 3 fixed at 0. Darker red indicates higher accuracy. Baseline *[0, 0, 0]* achieves 46.53%. Alignment-only training *[1, 0, 0]* decreases performance to 43.54%. (**B**) Performance comparison across configurations showing each stage's contribution. Error bars represent standard error. The full three-stage model *[1, 3, 3]* shows substantial improvement, with Stage 3 providing the largest gain (+13.33%, $p < 10^{-12}$). (**C**) Temporal evolution of model performance across training stages. Solid lines represent measurements, dashed lines show interpolated trajectories. (**D**) Optimization of Stage 3 duration using configuration *[1, 3, X]*. Performance peaks at X = 10 epochs (70.92%) before plateauing. (**E**) Effect of Stage 2 duration *[1, X, 10]*. Performance remains stable, with *[1, 10, 10]* achieving slightly better results (71.08%, p=0.9647). (**F**) Joint optimization of Stage 1 and Stage 2/3 durations (configurations [X, Y, Y]). Longer training shows improvements, leading to selection of *[5, 10, 10]* for final evaluation using both datasets.

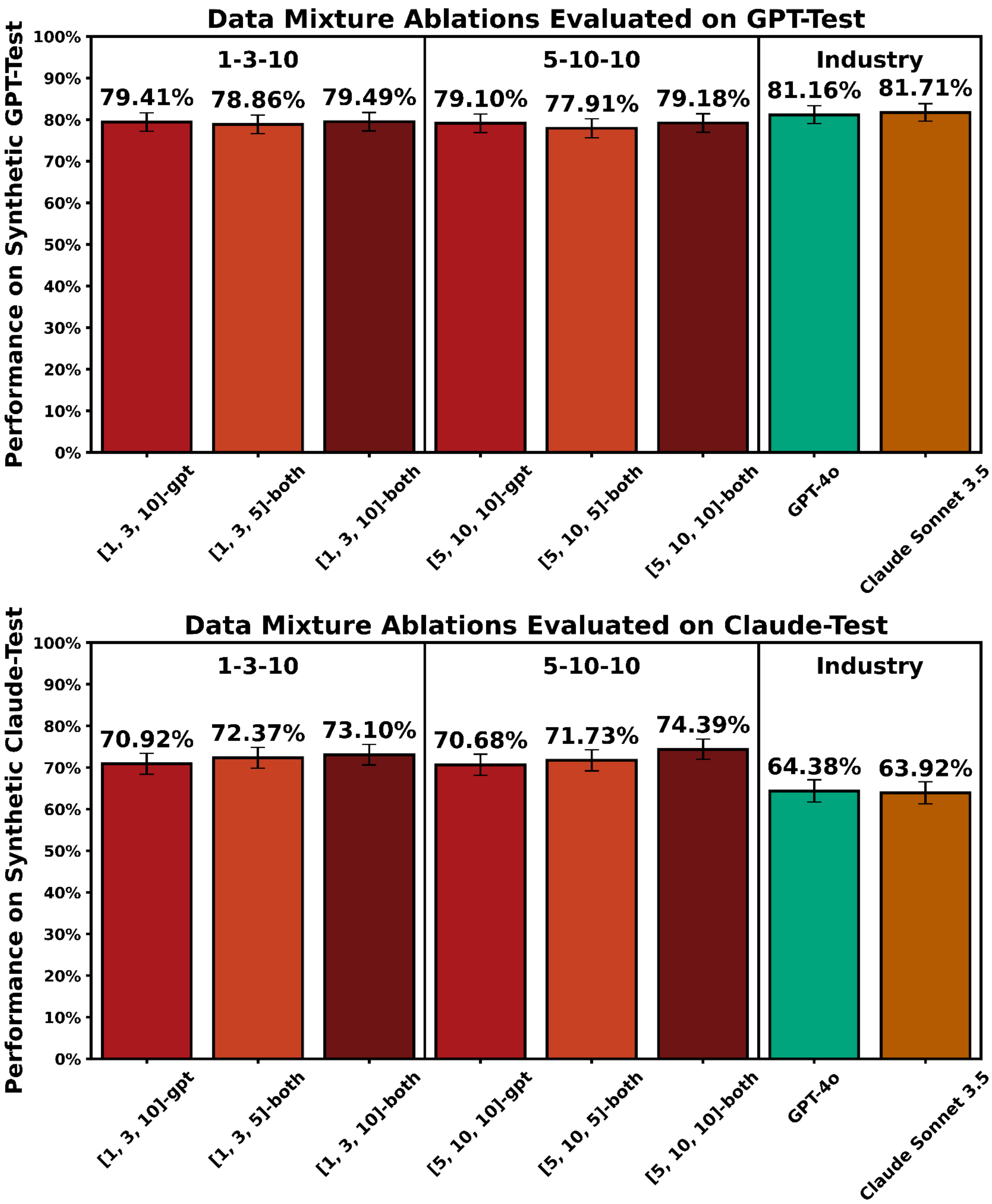

**Fig. S5. Ablation experiments for data mixtures.** (**A**) Performance on GPT generated MCQs (n=1,282) comparing models trained on GPT-generated data only (-*gpt*) versus both data sources (-*both*). Two base configurations *[1, 3, 10]* and *[5, 10, 10]* were each evaluated under two constraints when using both datasets: compute-matched (halved epochs) and epoch-matched (doubled compute due to doubled dataset). Including Claude-generated data showed no improvement over GPT-only training (p=1.000 for *[5, 10, 10]-gpt* vs. *[5, 10, 10]-both*). (**B**) Performance on Claude generated MCQs (n=1,239). Epoch-matched training on both datasets shows significant improvement (p=0.0427), with *[5, 10, 10]-both* reaching 74.39% compared to baseline Claude Sonnet 3.5, 63.92%. In both evaluations, epoch-matched consistently outperforms compute-matched training. Error bars represent standard error.

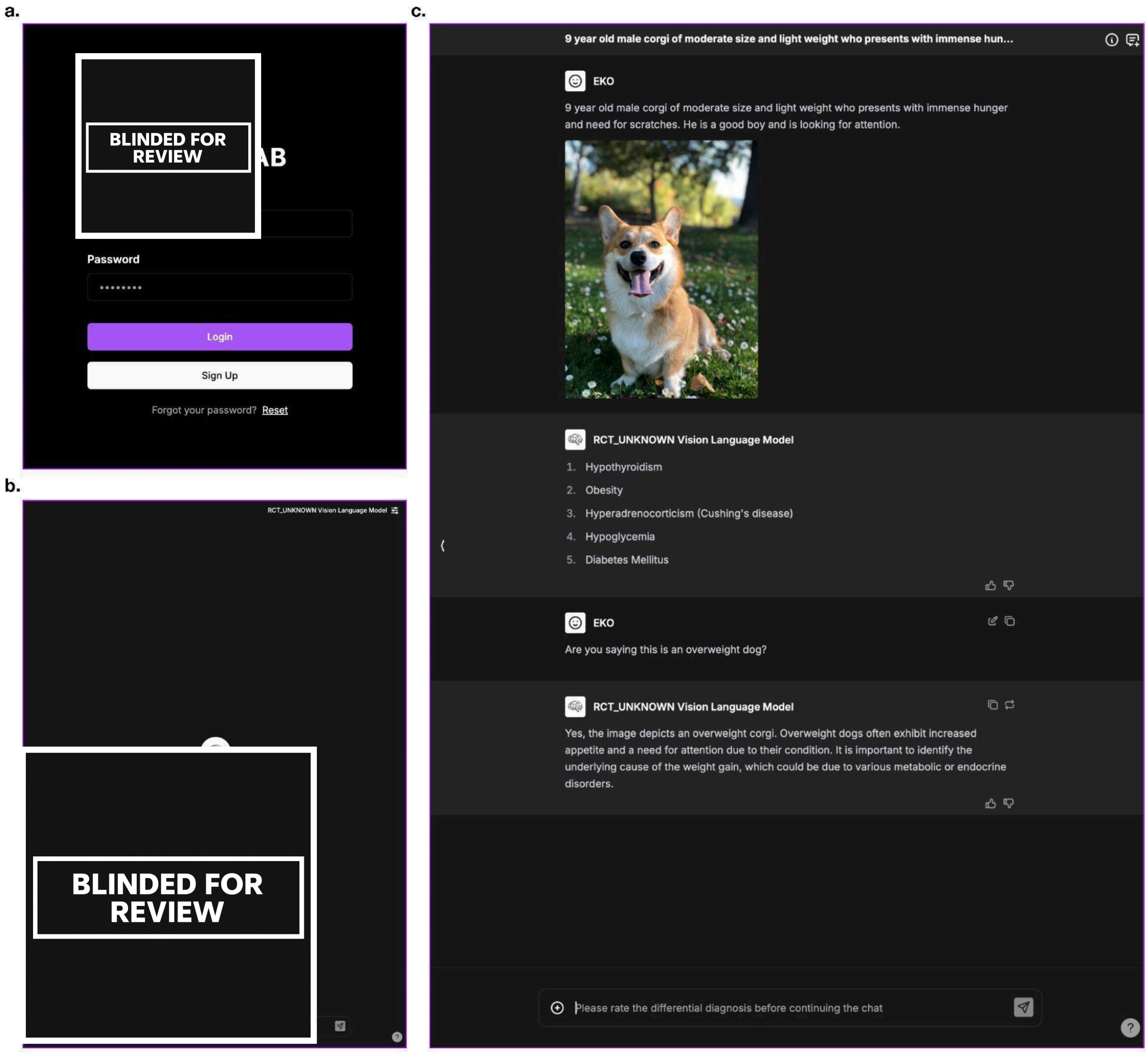

**Fig. S6. Randomized controlled trial user interface.** (**A**) The landing login interface. All participants had individual accounts with stored chat histories. Access was manually curated by the team. (**B**) Post-login, the participants are directly forwarded to the chatbot interface which is designed to mimic the typical chatbot interfaces, but enhanced with randomized back-end, a patient MRN field, and an ability to submit an image, including via intuitive “drag-and-drop”. (**C**) An example interaction with the model. The users can provide binary feedback for individual messages using “upvote” and “downvote” buttons. The model interacted with for this chat is CNS-Obsidian, but the participants are blinded to this information.

## Supplementary Methods

*Data Availability*

Data used for the general medicine stages of model training was downloaded from the public LLaVA-Med GitHub repository (https://github.com/microsoft/LLaVA-Med). Data from *Neurosurgery Publications* was used with permission of the publisher, but will not be released as it is protected by the respective copyrights and trademarks. *Self-Assessment of Neurological Surgery (SANS)* questions were used with permission of the *Congress of Neurological Surgeons (CNS)* and are intellectual property of the *CNS*. Clinical data was collected under a protocol approved by the NYU Langone Institutional Review Board (i23-00510) and is protected by the Health Insurance Portability Accountability Act. Icons were sourced from the Noun Project (https://thenounproject.com/).

*Code availability*

We used Python 3.10, 3.11, and 3.12 (*training*), as well as many open-source libraries, including datamapplot 0.4.2, HuggingFace Transformers 4.44.0, matplotlib 3.9.1, numpy 1.26.4, openai 1.55.3, pandas 2.2.0, pillow 10.1.0, pytorch 2.4, seaborn 0.13.2, wandb 0.17.6, among others. Our training and evaluations were executed using SLURM on the NYU Langone HPC cluster UltraViolet. We used NVIDIA Cuda 12.1. Our data filtering, data conversion, and training code will be publicly released on GitHub (https://github.com/alyakin314/CNS-Obsidian) upon publication of this work. The few-shot examples used in the conversion pipelines are omitted as they use excerpts from the *Neurosurgery Publications* materials. Weights for LLaVA-Next-Med-OLAB are publicly available (https://huggingface.co/NYU-OLAB/LLaVA-Next-Med-OLAB). CNS-Obsidian model will not be released as it was trained on proprietary content from Wolters Kluwer Wolters Kluwer, but can be made available to members of CNS upon request for research purposes.

*Data Acquisition*

We used AWS Textract to detect figures and captions locations and generate bounding boxes for cropping. A regex-based caption matcher then paired images with their corresponding captions using geometric distance calculations from a list of candidates. We extracted a total of 78,853 figures. We used Meta's Nougat9 to extract the remaining unstructured texts from the PDFs. We were able to extract a total of 139 million words. A regex-based system identified in-text figure mentions within the texts, linked them to corresponding figure captions, and stored the excerpts along with figure metadata.

*Image Filtering*

We built an image content classification system using a ResNet-50-based classifier[9] (feature extraction followed by a linear classifier) to categorize 500 manually annotated figures into three classes:

- Patient imaging (CT, MRI, X-ray, angiography)
- Clinical visuals (surgical fields, microscopy, anatomical drawings)
- Technical content (flowcharts, survival curves, tables)

We extracted the last layer of ResNet features from 400 images to train a linear classifier before validating its performance on the remaining 100 images (**Fig. S1**). We applied this model to filter the rest of the dataset. All conversion candidates underwent an initial filtering step, requiring >100 characters of associated text (combined caption and in-text mentions) for inclusion. For differential diagnosis dataset candidates, patient imaging was differentiated from everything else. For image-to-multiple choice question (MCQ) conversion candidates, patient imaging and clinical visuals were grouped and differentiated from technical content. For instruction fine-tuning dataset candidates, all three classes were differentiated from each other.

*Full Stack Web Application*

Randomization was performed independently for each patient encounter using Python's SciPy pseudorandomness. The user interface (**Fig. S6**) was adapted from a publicly available framework Chatbot UI, implemented in React and Next.js, and extended with features such as secure authentication, medical reference number recording, image submission, and endpoint randomization. The system utilized Postgres for account and chat storage, Flask with SQLiteDB for authentication, vLLM for hosting the local model (CNS-Obsidian), and Kong as an API gateway to connect with the PHI-safe OpenAI GPT-4o. The base version, CNS-Obsidian-Base [1, 3, 3], was used throughout the trial.

## Supplementary Results

### Ablation Experiments

*Notation*

We denote experimental configurations as [X, Y, Z], where X, Y, and Z represent the number of epochs in Stage 1 (alignment), Stage 2 (general fine-tuning), and Stage 3 (task-specific fine-tuning), respectively. Performance metrics are reported on two held-out test sets: GPT-Test (n=1,282 questions) and Claude-Test (n=1,239 questions). Due to the paper-level split we respected, no model has seen any figure from the same paper as any of the held out test questions. Most of our models are trained only on the data converted from the NeuroPubs using GPT-4o only. We refer to these models as [X, Y, Z]-gpt, or [X, Y, Z] for brevity. We also conducted ablations and experiments that used data converted via both GPT and Claude; those models will be explicitly referred to [X, Y, Z]-both.

*LLaVa-Next-Med Ablations*

We first validated our three-stage training approach on LLaVA-Next-Med by systematically varying Stage 1 and Stage 2 epochs while maintaining Stage 3 = 0. The baseline configuration [0, 0, 0] achieved 68.73% accuracy on GPT-Test and 46.53% on Claude-Test. Alignment-only training ([1, 0, 0]) decreased performance to 53.37% on GPT-Test and 43.54% on Claude-Test, indicating degradation of instruction-following capabilities. Stage 2 training improved performance, with [0, 3, 0] achieving 71.81% on GPT-Test and [3, 3, 0] reaching 54.60% on Claude-Test.

*CNS-Obsidian Missing Stages Ablations*

We then considered adding 3 epochs of Stage 3 to the [1, 3, 0] model (the length used by LLaVA-Med), making the [1, 3, 3] model, which served as our base CNS-Obsidian. To evaluate the contribution of each training stage, we conducted comprehensive ablation experiments on CNS-Obsidian by systematically removing different stages. We tested eight configurations: [0, 0, 0], [1, 0, 0], [0, 3, 0], [1, 3, 0], [0, 0, 3], [1, 0, 3], [0, 3, 3], and [1, 3, 3]. On GPT-Test the baseline [0, 0, 0] achieved 68.73%, while the full three-stage model [1, 3, 3] reached 76.41%, representing an 11.18% relative improvement. Inclusion of Stage 3, the neurosurgical specialization, leads to an improvement of +7,45% on GPT-Test ($p < 0.0001$) and +13.33% on Claude-test ($p < 10^{-12}$). The impact of Stage 2 (general medicine integration) was similarly present, but modest with +2.38% (p=0.1848) and +3.40% (p=0.0522) improvements on GPT-Test and Claude-Test, respectively. Lastly, the inclusion of Stage 1 alignment did not affect performance on GPT-Test, but boosted the performance on Claude-Test by +1.54% (p=0.3994). These results confirm the complementary nature of all training stages.

*Optimization of Task-Specific Fine-Tuning*

We investigated the optimal duration of task-specific fine-tuning (Stage 3) by evaluating configurations [1, 3, X] where $X \in \{0, 1, 3, 5, 10, 20\}$. Performance exhibited a monotonic improvement until X = 10, with the model achieving peak performance on both GPT-Test (79.48%) and Claude-Test (70.92%). Training beyond 10 epochs led to performance plateauing and slight decreases.

*General Fine-Tuning Duration Analysis*

To optimize Stage 2, we evaluated configurations [1, X, 10] where $X \in \{3, 5, 10\}$. Remarkably, performance remained stable across different Stage 2 durations. For GPT-Test the best-performing checkpoint of these was [1, 3, 10] at 79.41%. For Claude-Test the [1, 10, 10] slightly outperformed the other two models at 71.08%. Notably, the differences between the two were not statistically significant on both GPT-Test (p=0.8079) and Claude-Test (p=0.9647). We decided to use [1, 3, 10] as one of the checkpoints to retrain using both datasets, but also to evaluate different training lengths of checkpoint that have the same number of Stages 2 and 3.

*Comprehensive Stage Length Optimization*

Our final experimental series explored the joint optimization of alignment and fine-tuning stages through configurations [X, Y, Y] where $X \in \{1, 3, 5\}$ and $Y \in \{3, 5, 10\}$. Overall, we observed an expected pattern of improvements with longer fine-tuning stages. The changes between the alignment length were not significant, but generally leaned towards better performance with longer training, at least on GPT-Synth, so we picked [5, 10, 10] as the other candidate to be retrained with both GPT and on both datasets.

*Secondary Data Source*

To maximize model performance, we evaluated our top configurations ([1, 3, 10] and [5, 10, 10]) using data processed by both GPT and Claude. [5, 10, 10]-both achieved 79.18% on GPT-Test and 74.39% on Claude-Test. For comparison, GPT-4o achieved 81.16% on GPT-Test and Claude Sonnet 3.5 reached 63.92% on Claude-Test.